\documentclass[12pt]{article}

\usepackage{authblk}
\usepackage{cite}
\usepackage{amsmath,amssymb,amsfonts}
\usepackage{algorithmic}
\usepackage{graphicx}
\usepackage{textcomp}
\usepackage{xcolor}
\usepackage{url}
\usepackage{hyperref}
\usepackage{comment}
\usepackage[margin=1in]{geometry}
\def\BibTeX{{\rm B\kern-.05em{\sc i\kern-.025em b}\kern-.08em
    T\kern-.1667em\lower.7ex\hbox{E}\kern-.125emX}}
\begin{document}

\title{Explainable Face Presentation Attack Detection via Ensemble-CAM}

\newcommand{\clrksn}{{*}}
\newcommand{\rit}{{$^\mathsection$}}
\newcommand{\afrl}{{$^{\ddagger}$}}

\renewcommand*{\Authsep}{\ }
\renewcommand*{\Authand}{\ }
\renewcommand*{\Authands}{\ }

\author[\clrksn]{Rashik Shadman}
\author[\afrl]{M G Sarwar Murshed} 
\author[\clrksn]{Faraz Hussain}

\affil[\clrksn]{Clarkson University, Potsdam, NY, USA\authorcr {\tt \{shadmar,fhussain\}@clarkson.edu}\vspace{0.4em}}
\affil[\afrl]{University of Wisconsin-Green Bay, Green Bay, WI, USA\authorcr{\tt murshedm@uwgb.edu}\vspace{0.4em}}

\maketitle

\begin{abstract}
Presentation attacks represent a critical security threat where adversaries use fake biometric data — such as face, fingerprint, or iris images — to gain unauthorized access to protected systems. Various presentation attack detection (PAD) systems have been designed leveraging deep learning (DL) models to mitigate this type of threat. 
Despite their effectiveness, most of the DL models function as black boxes - their decisions are opaque to their users. The purpose of explainability techniques is to provide detailed information about the reason behind the behavior/decision of DL models. In particular, visual explanation is necessary to better understand the decisions/predictions of DL-based PAD systems and determine the key regions due to which a biometric image is considered real or fake by the system. In this work, a novel technique, Ensemble-CAM, is proposed for providing visual explanations for the decisions made by deep learning-based face PAD systems. Our goal is to improve DL-based face PAD systems by providing a better understanding of their behavior. Our provided visual explanations will enhance the transparency and trustworthiness of DL-based face PAD systems.
\end{abstract}

\maketitle

\section{Introduction}
Recently, deep learning models have been adopted for PAD systems \cite{bhattacharjee2019recent}. The field of presentation attack detection has significantly advanced using deep learning models. DL models offer highly effective techniques to detect fraudulent attempts. These models learn complicated patterns and features by leveraging large datasets and complex neural network architectures. Thus, the accuracy and robustness of PAD systems are enhanced. 

A major drawback of AI is that DL models act in a black-box manner and lack transparency \cite{R0}. Despite the high accuracy rates of artificial neural networks, it is important to understand their decisions and reasoning. Therefore, explainability techniques are necessary to explain the reasons behind the predictions of DL models. Specifically, presentation attack detection may be significantly impacted by using explainability techniques. The purpose of PAD systems is to detect fake biometric traits. In the case of a DL-based PAD system, the system's predictions will be more convincing and trustworthy with proper explanations (the rationale behind the predictions).

Sequeira et al. \cite{sequeira2021exploratory} used the Grad-CAM method to generate the explanations for face presentation attack detection. They considered two different evaluation frameworks to compute the variability of the explanations for both genuine and imposter samples. Huber et al. \cite{huber2023explainability} used Grad-CAM and Grad-CAM++ to explain the behavior of face PAD and investigate the gender bias of the generated explanations.


This research adopts a new method to explain DL-based face presentation attack detection visually. This method is an ensemble of gradient-based discriminative localization methods, viz. Grad-CAM \cite{R5}, HiResCAM \cite{draelos2020use}, and Grad-CAM++ \cite{chattopadhay2018grad}. The goal is to perform improved and very narrow/specific localization of the most relevant regions of genuine/fake face images using the Ensemble-CAM method. 
The most significant regions of a fake face image (the regions that are different from a genuine face image) can be identified very accurately (compared to other gradient-based localization methods) by this new Ensemble-CAM method. Very narrow/specific localization highlights the most important features (important to the DL model for its prediction) of a face image.

The main contributions of this paper are:
\begin{itemize}
    \item Development of an efficient deep learning-based PAD model to detect genuine/fake face images. A publicly available image-based face presentation attack dataset is used to train and test the deep learning model. The test accuracy of the DL model is computed.
    \item Application of a novel method, Ensemble-CAM, to provide visual explanations for the predictions/decisions of the DL-based face PAD system. This method can be used to visually explain the results of the existing face PAD models as well as the newly developed face PAD model. The provided visual explanations will increase the transparency of DL-based face PAD systems and help detect vulnerabilities in face PAD systems.
    \item Evaluation of the proposed Ensemble-CAM method and comparing this novel method with Grad-CAM, HiResCAM, and Grad-CAM++.
    \item The code is  available at: \newline \url{https://github.com/rashikshadman/Ensemble-CAM}.
\end{itemize}

The remainder of this paper is organized as follows. Section \ref{RW} reviews the related work. Section \ref{PA} defines presentation attacks. Section \ref{M} outlines the proposed methodology, and Section \ref{EC} presents the algorithm. Section \ref{model} provides details of the model used. Section \ref{Visual} illustrates visual explanations generated by the proposed Ensemble-CAM method. Section \ref{Evaluation} reports the evaluation results, followed by a discussion in Section \ref{Discuss}. Finally, Section \ref{end} concludes the paper.

\section{Related Work}\label{RW}
In this section, we describe the CAM, Grad-CAM, HiResCAM, and Grad-CAM++ techniques, which are prominent explainability tools.

Several recent works have leveraged Grad-CAM or similar gradient-based visualization techniques to explain the decision-making process of face presentation attack detection (PAD) models. Muhammad et al. \cite{muhammad2023self} applied Grad-CAM and LIME to a self-supervised PAD framework to visualize how the DGS mechanism improves PAD by directing the model’s focus toward meaningful cues such as paper artifacts, device edges, and motion patterns. These focused attention regions help the model rely on intrinsic features rather than broad image areas, enhancing its ability to distinguish real from spoofed faces. Pan et al. \cite{pan2022attention} propose an explainable face PAD framework that generates both visual and verbal explanations using Grad-CAM saliency maps and LSTM gradients. By leveraging spatial and temporal information, the approach highlights spoof-related anomalies and enhances classification performance. These studies show that gradient-based visualization methods are powerful tools for verifying model behavior, ensuring that PAD systems rely on semantically relevant features, and enhancing user trust in biometric security applications.

Class Activation Mapping (CAM) \cite{R4} is a very effective visual explainability technique. A class activation map of a specific class serves to delineate the discerning regions within an image that a convolutional neural network (CNN) utilizes in the identification of said class. The architecture predominantly comprises convolutional layers, and in the proximity of the final output layer, specifically the softmax layer in categorical classification scenarios, a global average pooling operation is conducted on the convolutional feature maps. These pooled features subsequently serve as input to a fully connected layer, responsible for generating the ultimate output, be it categorical or otherwise.
The discernment of the relative importance of image regions is achieved by back-projecting the weights of the output layer onto the convolutional feature maps. The resulting CAM effectively highlights the regions within the image that are discriminative and class-specific.

Gradient-weighted Class Activation Mapping (Grad-CAM) \cite{R5} is a generalization of CAM. Grad-CAM leverages the gradients associated with a designated target concept, such as the logits corresponding to the classification of `dog' or even a descriptive caption. These gradients are traced back through the network to the final convolutional layer, facilitating the generation of a coarse localization map.
This map effectively highlights crucial regions within the image that play a significant role in predicting the specified concept. The neurons within the ultimate convolutional layers specifically seek semantic, class-specific information within the image, such as distinct object parts. By utilizing the gradient information streaming into the last convolutional layer of the CNN, Grad-CAM discerns the significance of each neuron in relation to a particular decision of interest. This method is notably characterized by its high degree of class discrimination. It uses the gradients of any target concept (such as logits for `dog'), flowing into the final convolutional layer to generate a coarse localization map that highlights the important regions in the input image for predicting the concept.

\begin{figure*} [hbt!]
\center
\includegraphics[width=\linewidth]{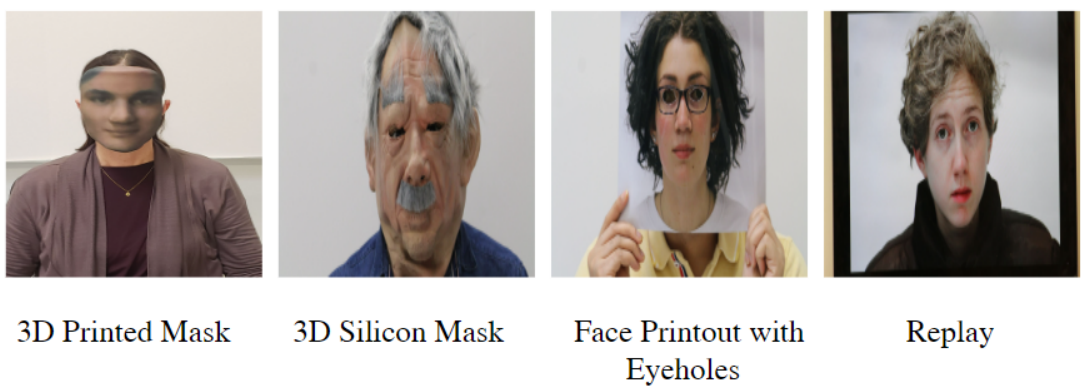}
  \caption{Examples of presentation attacks on face recognition systems, such as 3D mask (printed and silicon), face printout (with eyeholes), replay attack (mobile display) \cite{purnapatra2021face}.} 
  \label{fig:PA_1}
\end{figure*}

Draelos and Carin \cite{draelos2020use} analyzed the limitations of the Grad-CAM method. In some cases, Grad-CAM highlights irrelevant regions (the model did not use those regions) as a side effect of the gradient averaging step. Draelos and Carin proposed a novel explanation method named HiResCAM. This class-specific method ensures the localization of only the relevant regions that the model utilizes for its prediction. Draelos and Carin performed experiments to prove that HiResCAM's explanations more accurately reflect the model than Grad-CAM's. The main difference between Grad-CAM and HiResCAM: Grad-CAM uses the average gradient and multiplies it with the feature maps; HiResCAM performs element-wise multiplication of the gradient and the feature maps. The advantages of HiResCAM over Grad-CAM are described below \cite{draelos2020use}.
\begin{itemize}
    \item HiResCAM explanations are often more focal than those of Grad-CAM.
    \item In some cases, HiResCAM localizes the correct object while Grad-CAM cannot.
    \item More relevant areas of the image are highlighted by HiResCAM than Grad-CAM.
\end{itemize}

Chattopadhay et al. \cite{chattopadhay2018grad} proposed a new method named Grad-CAM++, built on Grad-CAM. They aimed to generate better visual explanations of CNN model predictions using the Grad-CAM++ method. Chattopadhay et al. presented two shortcomings of the Grad-CAM method, which they tried to overcome by the Grad-CAM++ method: 1) failure to localize multiple occurrences of the same class in an image, 2) failure to highlight the full region of the class. Grad-CAM++ is a generalization of Grad-CAM. The main difference is that Grad-CAM++ uses second order gradients.

The proposed Ensemble-CAM method combines these three described methods to localize the most important features relevant to the DL model for predicting genuine/fake face images. The Ensemble-CAM method can perform very narrow/specific localization of the most significant regions of the input image.

\begin{figure*} [hbt!]
\center
\includegraphics[width=\linewidth]{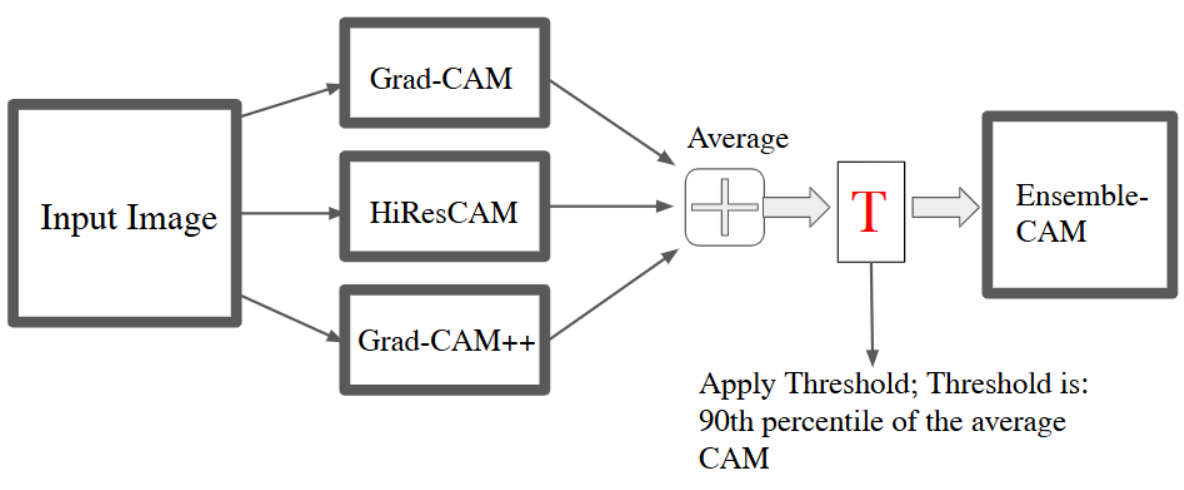}
  \caption{An overview of the Ensemble-CAM method. For the input image, three CAMs are generated using the Grad-CAM, HiResCAM, and Grad-CAM++ methods. Then, an average CAM is computed to combine all the features of these three CAMs. The Ensemble-CAM is generated by applying a threshold to the average CAM.} 
  \label{fig:ensemble}
\end{figure*}

\section{Presentation Attacks}\label{PA}
Biometric systems are extensively used to ensure the security and privacy of users. Unique biological characteristics include face, fingerprint, or iris patterns. Biometric systems use these traits to differentiate between genuine individuals and impostors. Presentation or spoofing attacks use fake biometric data to deceive biometric security systems. These attacks are a threat to security and authentication and need to be mitigated. Husseis et al. \cite{husseis2019survey} presented a comprehensive analysis of presentation attack and presentation attack detection.

Face recognition faces different presentation attacks, such as:
\begin{itemize}
    \item Printed face image
    \item 3D printed mask \& 3D silicon mask \cite{erdogmus2013spoofing}, \cite{erdogmus2014spoofing}
    \item Display image \& Display video \cite{maatta2011face}, \cite{patel2015live}
    \item Facial accessories \cite{min2011improving}
    \item Artificial and natural facial hair \cite{singh2019recognizing}, \cite{dhamecha2014recognizing}
    \item Facial makeup \cite{zheng2017multi}  
\end{itemize}
Fig. \ref{fig:PA_1} shows examples of potential face presentation attacks.

\section{Methodology}\label{M}
In this section, we explain the Grad-CAM \cite{R5}, HiResCAM \cite{draelos2020use}, and Grad-CAM++ \cite{chattopadhay2018grad} methods in detail. Our method, Ensemble-CAM, is a combination of these three methods. 

Grad-CAM is a generalization of CAM. In CNN, the class score $Y^c$ for class c is backpropagated till the last convolutional layer. The gradient of the class score $Y^c$ is computed with respect to feature maps $A^k$ of a convolutional layer, i.e., $\frac{\partial {Y^c}}{\partial {A^k}}$. Then the weight $w_{k}^{c}$ is computed for feature map $k$ and a target class $c$. While computing the weight $w_{k}^{c}$, Grad-CAM uses global average pooling. The details are described in \cite{R5}.

\begin{equation}
   w^c_k = \frac{1}{Z}\sum_{i}\sum_{j}\frac{\partial {Y^c}}{\partial {A^k_{ij}}} 
\end{equation}

On the other hand, Grad-CAM++ takes a weighted combination of positive partial derivatives to compute the weight $w_{k}^{c}$ \cite{chattopadhay2018grad}.

\begin{equation}
   w^c_k = \sum_{i}\sum_{j}\alpha_{ij}^{kc}.relu(\frac{\partial {Y^c}}{\partial {A^k_{ij}}}) 
\end{equation}

Finally, a linear combination of the forward activation maps followed by a \emph{relu} layer gives the final class activation map \cite{chattopadhay2018grad}.

\begin{equation}
    L^c_{ij} = relu(\sum_{k}w^c_k.A^k_{ij})
\end{equation}

The first step of HiResCAM is computing the gradient of class score $Y^c$ with respect to the feature maps $A^k$. Then, element-wise multiplication of the gradient and the feature maps is performed to generate the class activation map \cite{draelos2020use}.

\begin{equation}
    L^c = \sum_{k}\frac{\partial {Y^c}}{\partial {A^k}}A^k
\end{equation}

\begin{table}[ht]
\caption{Face PAD dataset describing the number of images for each class used for training, validation, and testing.}
\label{face_pad}
\centering
\begin{tabular}{|c|c|c|}
\hline
  & \textbf{Spoof} & \textbf{Live}\\
\hline
\textbf{Train} & 65534 & 65534\\
\hline
\textbf{Validation} & 2000 & 2000\\
\hline
\textbf{Test} & 2000 & 2000\\
\hline
\end{tabular}
\end{table}

\section{Ensemble-CAM}\label{EC}

The proposed Ensemble-CAM approach integrates three widely used class activation mapping techniques: Grad-CAM, HiResCAM, and Grad-CAM++, to generate a more robust and informative visualization of model interpretability. An overview of the Ensemble-CAM framework is presented in Fig. \ref{fig:ensemble}. Initially, the input image is processed by all three CAM algorithms individually, producing three distinct saliency maps that capture different aspects of the model’s feature importance. These individual CAMs are then aggregated by computing their pixel-wise average, as shown in equation \ref{eq1}. The resulting average CAM incorporates the complementary feature representations contributed by each of the three methods, thereby enhancing the overall interpretability and robustness of the explanation.

To further refine the visualization and emphasize the most salient regions, a thresholding operation is applied to the average CAM (equation \ref{eq2}). Specifically, the threshold is chosen as the 90th percentile of the pixel intensity distribution of the average CAM. This percentile-based approach ensures that the top 10\% of the most discriminative features are retained, while less informative regions are suppressed by setting their values to zero. The final Ensemble-CAM thus highlights the most critical features shared across Grad-CAM, HiResCAM, and Grad-CAM++, producing a more precise and reliable explanation of the model’s decision-making process.

In our implementation, we build upon the open-source CAM library by Gildenblat et al. \cite{jacobgilpytorchcam} and extend its functionality to generate the Ensemble-CAM visualizations.

\begin{equation}\label{eq1}
Avg\_CAM = \frac{Grad\_CAM + HiResCAM + Grad\_CAM++}{3}
\end{equation}

\begin{equation}\label{eq2}
Ensemble\_CAM = Apply\_Threshold(Avg\_CAM)
\end{equation}

\begin{figure*}[ht]
    \center
    \includegraphics[width=\linewidth]{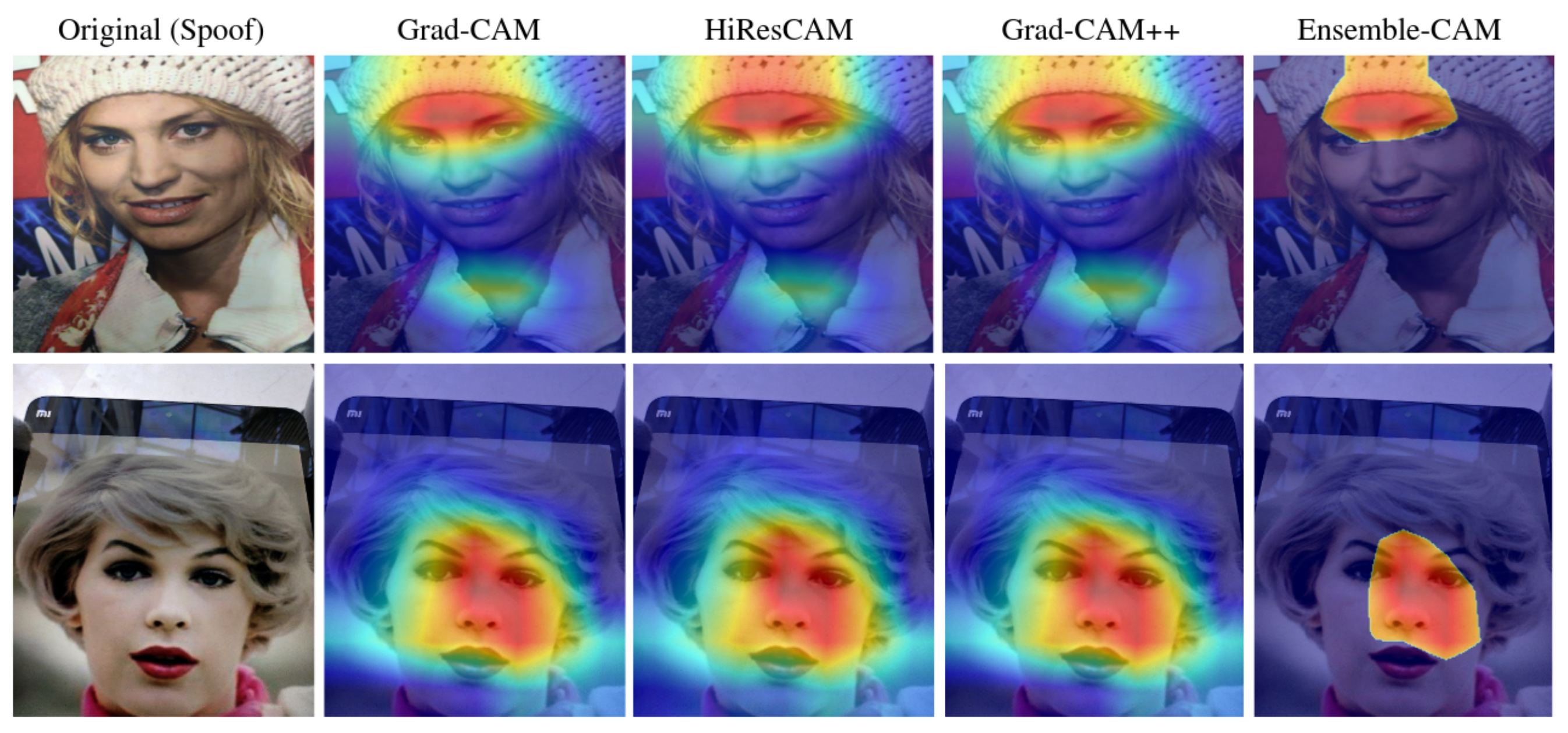} 
    \caption{Visual explanations of face PAD model predictions. Here, two spoof test images are considered. The model predicts that these are spoof images (correct prediction). From the left, the first image is the original spoof image. Grad-CAM, HiResCAM, and Grad-CAM++ are generated using the PAD model for the predicted class and overlaid on the test image. The red, yellow, and green regions highlight the relevant features, while the blue regions highlight the non-relevant features. The last image in the row shows the Ensemble-CAM overlaid on the test image, highlighting the most important features (for the correct prediction by the model) very specifically and accurately.}
    \label{fig:top}
\end{figure*}

\begin{figure*}[ht]
    \includegraphics[width=\linewidth]{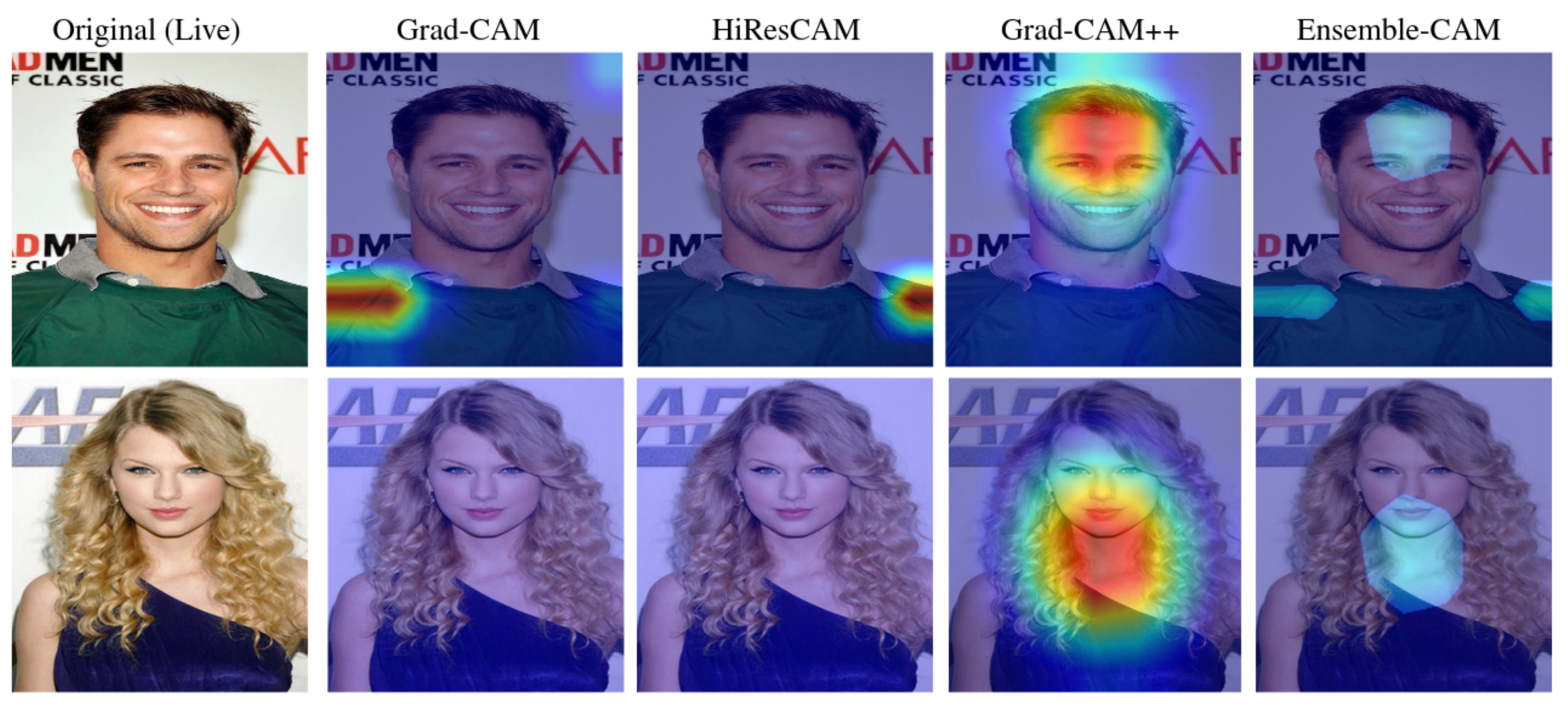}
    \caption{Visual explanations of face PAD model predictions. Here, two live test images are considered. The model predicts that these are live images (correct prediction). From the left, the first image is the original live image. Grad-CAM, HiResCAM, and Grad-CAM++ are generated using the PAD model for the predicted class and overlaid on the test image. The red, yellow, and green regions highlight the relevant features, while the blue regions highlight the non-relevant features. The last image in the row shows the Ensemble-CAM overlaid on the test image, highlighting the most important features (for the correct prediction by the model) very specifically and accurately.}
    \label{fig:bottom}
\end{figure*}

\section{Face PAD Model}\label{model}
We use a DenseNet-161 model for face presentation attack detection. The model is pre-trained on the ImageNet dataset \cite{deng2009imagenet}. The pre-trained DenseNet-161 model is fine-tuned on a subset of the CelebA-Spoof face PAD dataset \cite{zhang2020celeba} (a public dataset). The CelebA-Spoof dataset is a large-scale, richly annotated benchmark for face anti-spoofing, featuring diverse spoof types, facial attributes, and real-world variations. Its extensive size and fine-grained labels make it highly suitable for training and evaluating robust, generalizable face presentation attack detection models. The description of the dataset is shown in Table \ref{face_pad}. The AdamW optimizer is used for fine-tuning the model. The learning rate is 0.0005, and the number of epochs is 20. StepLR learning rate scheduler is used where step\_size = 7 and gamma = 0.1. 

For the test set, the model's APCER (Attack presentation classification error rate) is 12.4\%, and BPCER (Bona Fide presentation classification error rate) is 0.95\%. The model's overall test accuracy is 93.33\%.

\begin{figure*}[ht]
\centering
\includegraphics[width=\linewidth]{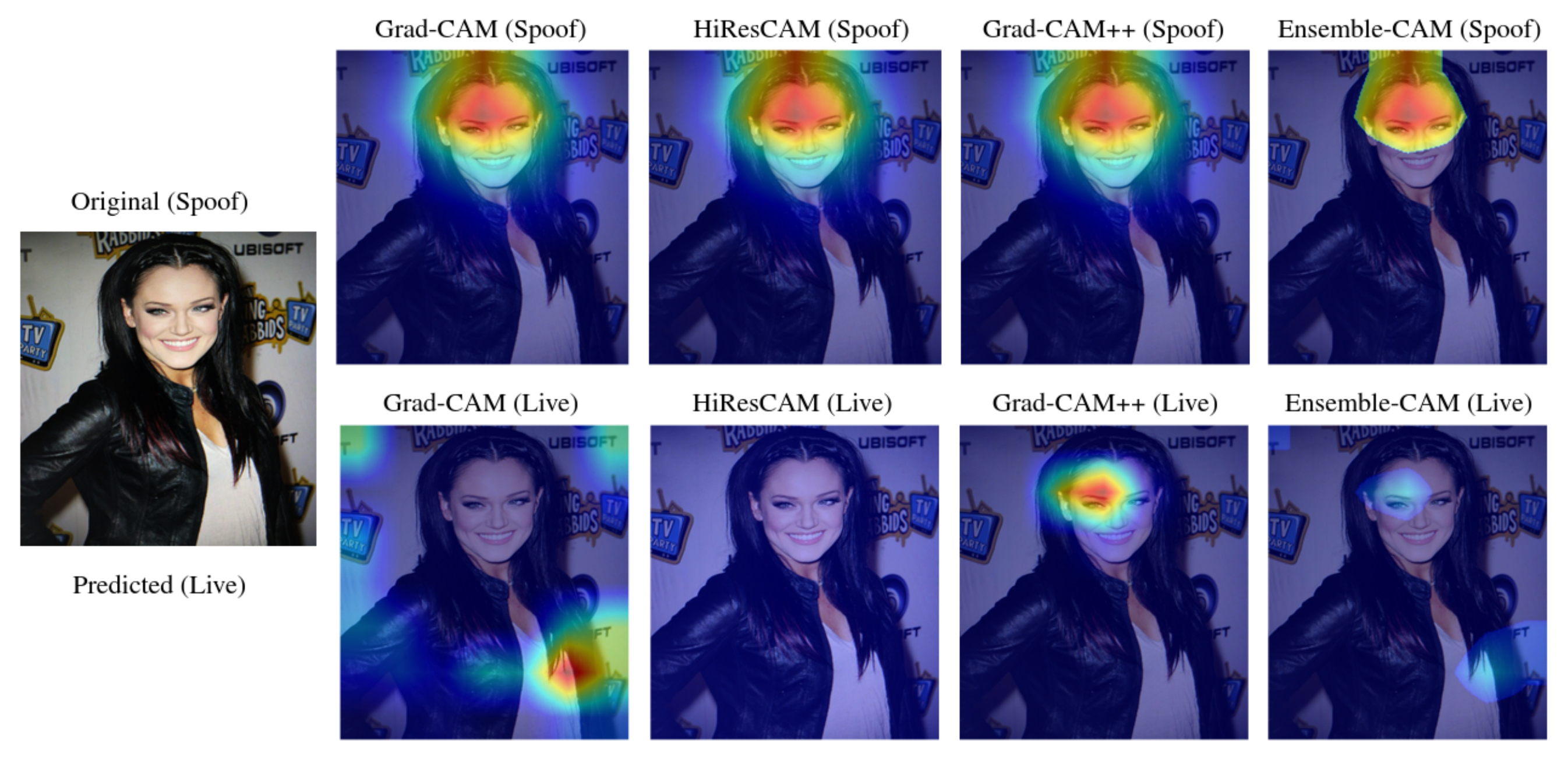}
\caption{Visual explanation of wrong prediction made by the face PAD model. Here, a spoof test image is considered. The model predicts that this is a live image (wrong prediction). The top row shows the CAMs for the original class (spoof), and the bottom row shows the CAMs for the predicted class (live). The Ensemble-CAM of the original class highlights the upper part of the face (nose, eyes, and forehead). The Ensemble-CAM of the wrongly predicted class highlights the right eye, right forehead, and the left side of the body; the model fails and makes a wrong prediction due to these features.}
\label{fig:wrong}
\end{figure*}

\section{Visual Explanations of Face PAD Model Decisions}\label{Visual}
In this section, visual explanations are generated for the decisions of the face PAD model using the Ensemble-CAM method and other gradient-based methods for analysis. 

In Fig. \ref{fig:top} and Fig. \ref{fig:bottom}, Ensemble-CAM results are shown for four test images (two spoof images and two live images). Also, Grad-CAM, HiResCAM, and Grad-CAM++ results are shown for comparison. All the CAMs are generated using the face PAD model for predicted classes (by the model). For a test image, all the generated CAMs are overlaid on it.

In Fig. \ref{fig:top}, there is not much difference between Grad-CAM, HiResCAM, and Grad-CAM++ results. However, in Fig. \ref{fig:bottom}, Grad-CAM, HiResCAM, and Grad-CAM++ results are fully different. This shows that Grad-CAM, HiResCAM, and Grad-CAM++ results are not always the same. The Ensemble-CAM combines the most important features of Grad-CAM, HiResCAM, and Grad-CAM++. The Ensemble-CAM performs very narrow/specific localization of the most relevant features, which the PAD model considers for its prediction.

In the top row of Fig. \ref{fig:top}, the most important region of the test spoof image is \emph{the middle of the forehead} (highlighted in the Ensemble-CAM image). In the bottom row, the Ensemble-CAM highlights \emph{the nose and the area around the nose} of the test image.

In the top row of Fig. \ref{fig:bottom}, the Ensemble-CAM highlights the three most important regions of the test live image. Among these three regions, two of them are not on the face and one of them is \emph{the middle of the forehead}. In the bottom row, the most important region is \emph{the lips and the neck} shown in the Ensemble-CAM image.

In Fig. \ref{fig:wrong}, Ensemble-CAM results are shown for a wrongly predicted test image. Originally, the test image is of a spoof face, but the model predicts it as a live face. Also, Grad-CAM, HiResCAM, and Grad-CAM++ results are shown for comparison. All the CAMs are generated using the face PAD model and overlaid on the test image. The top row shows the CAMs generated for the original class (spoof), and the bottom row shows the CAMs generated for the predicted class (live). The Ensemble-CAM of the live class highlights the features for which the model makes the wrong prediction.

The Ensemble-CAM results show that the Ensemble-CAM combines the most important features of Grad-CAM, HiResCAM, and Grad-CAM++; and performs very narrow/specific localization of the most relevant regions of the spoof/live face image, which the model considers for its prediction. This way, the Ensemble-CAM performs better than other gradient-based methods because it covers the flaws of Grad-CAM, HiResCAM, and Grad-CAM++ methods.

\section{Evaluation of the Ensemble-CAM Results}\label{Evaluation}
Here, we assess the visual explanations produced by the Ensemble-CAM method for the face PAD model using the retention method \cite{chattopadhay2018grad}. 
The retention method measures how well the explanation preserves the model’s confidence when only the most important regions are retained — a smaller drop suggests more effective identification of key features.

\begin{figure*}[ht]
    \centering
    \includegraphics[width=\linewidth]{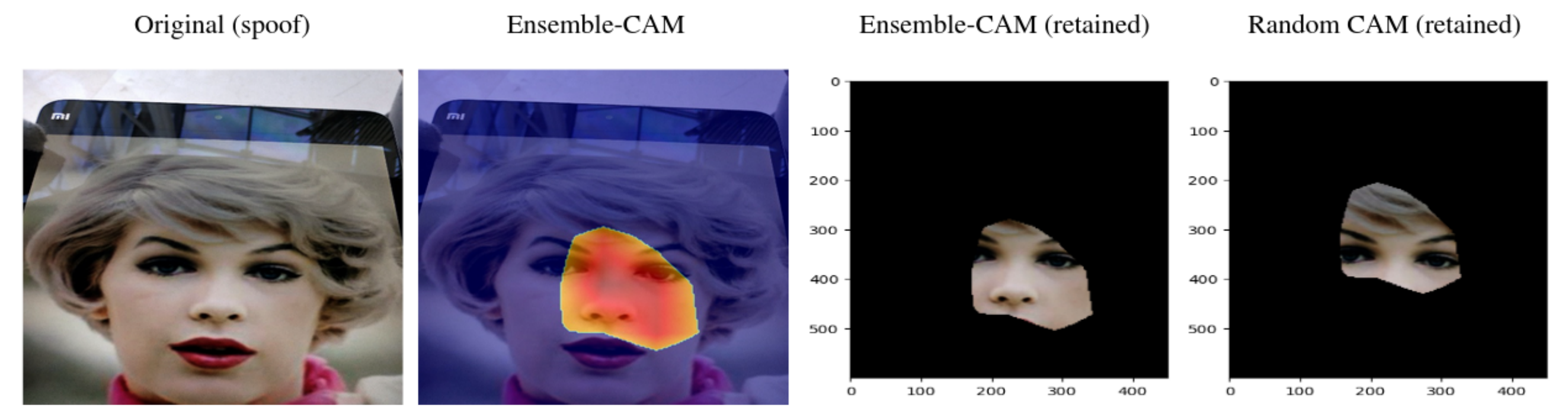} 
    \caption{An overview of the retention scheme used to evaluate the visual explanations generated by the Ensemble-CAM method is presented. From left to right, the first image shows the original image, the second image displays the Ensemble-CAM overlaid on the image, the third image retains the Ensemble-CAM regions, and the last image retains random CAM regions of the same dimension.}
    \label{fig:rtn_random}
\end{figure*}

\begin{table*}[htbp!]
\caption{Performance Comparison between the Ensemble-CAM and random CAM using the retention method. The results are generated for the whole test dataset.}
\label{tabel_rtn_random}
\centering
\resizebox{\linewidth}{!}{%
\begin{tabular}{|c|c|c|c|c|}
\hline
  \textbf{Metrics} & \textbf{Ensemble-CAM} & \textbf{Random CAM}\\
\hline
\textbf{Average Confidence Drop (lower is better)} & 15.43\% & 26.42\%\\
\hline
\textbf{Prediction Change Percentage (lower is better)} & 15.90\% & 26.90\%\\
\hline
\end{tabular}}
\end{table*}

\begin{figure*}[ht]
    \centering
    \includegraphics[width=\linewidth]{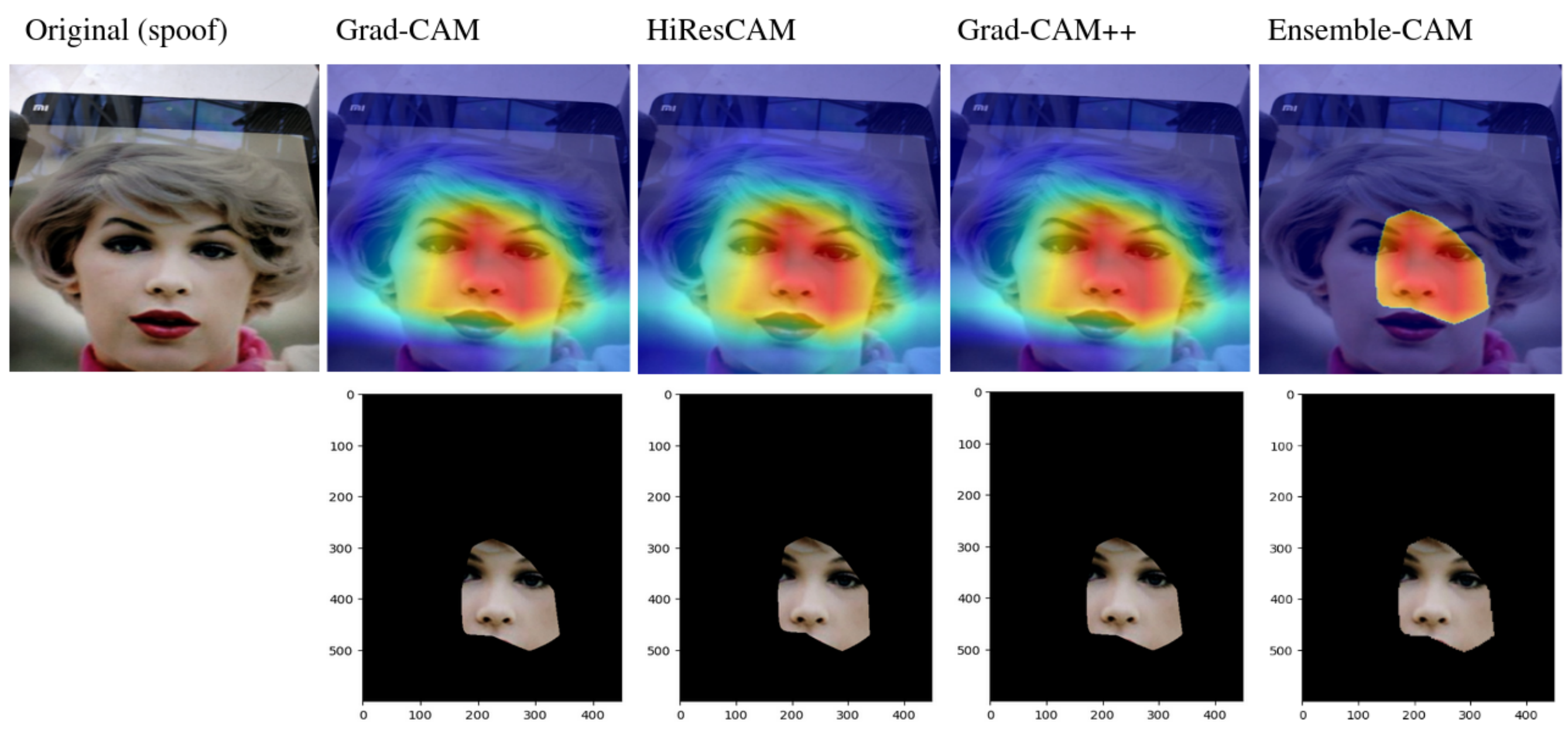} 
    \caption{Comparison of Ensemble-CAM explanation with that of other gradient-based methods using the retention scheme. The top row shows the original test image, and Grad-CAM, HiResCAM, Grad-CAM++, and Ensemble-CAM are overlaid on the test image. In the bottom row, the top 10\% most important regions identified by Grad-CAM, HiResCAM, and Grad-CAM++ are retained, and in the last image, Ensemble-CAM regions are retained. We retain the top 10\% most important regions highlighted by other CAMs except Ensemble-CAM to keep the size of the retained area the same for the comparison.}
    \label{fig:rtn}
\end{figure*}

The Ensemble-CAM highlights the most relevant features by combining the outputs of Grad-CAM, HiResCAM, and Grad-CAM++ methods. A threshold is set at the 90th percentile of the average CAM to retain the top 10\% of the most significant values. All CAM values below this threshold are set to zero, effectively filtering out less important regions and preserving only the most informative features.

During the retention process, we retain only the Ensemble-CAM regions, and other regions of the image are removed/covered. Fig. \ref{fig:rtn_random} shows an example of the retention method. To enable a fair comparison with the Ensemble-CAM regions, random areas of identical dimensions are retained, referred to here as the random CAM. 

When the most important regions of a test image are retained, the prediction confidence of the model typically experiences a drop, but it will not be as significant. The more important the retained regions, the smaller the drop in confidence.
For instance, suppose the model initially predicts the correct class with 99\% confidence. After retaining the important CAM regions, the confidence drops to 80\%, resulting in a 19\% reduction. This is an example, and the reduction in confidence varies for each CAM method. Another evaluation metric is the prediction change rate — the percentage of test images for which the model's predicted class changes after retention. Like the confidence drop, the lower this rate, the better in the case of the retention method.

When the Ensemble-CAM regions are retained, the confidence drop should be lower, as the Ensemble-CAM regions are the most relevant regions to the model. On the other hand, when random regions are retained, the confidence drop should be higher, as these regions are not that significant in general. Also, the prediction change percentage should be lower for the Ensemble-CAM than for the random CAM. For the entire test dataset (comprising 4000 images), the average confidence drop for the retention method is 15.43\% for Ensemble-CAMs. For random CAMs, the average confidence drop is 26.42\%, which is higher compared to the average drop for Ensemble-CAMs. The prediction change percentage for Ensemble-CAMs is 15.90\%. For random CAMs, the prediction change percentage is 26.90\%. The results are shown in Table \ref{tabel_rtn_random}.

\begin{table*}[ht]
\caption{Performance Comparison of Ensemble-CAM with Grad-CAM, HiResCAM, and Grad-CAM++ using the retention method. The results are generated for the whole test dataset.}
\label{tabel_rtn}
\centering
\resizebox{\linewidth}{!}{%
\begin{tabular}{|c|c|c|c|c|}
\hline
  \textbf{Metrics} & \textbf{Grad-CAM} & \textbf{HiResCAM} & \textbf{Grad-CAM++} & \textbf{Ensemble-CAM}\\
\hline
\textbf{Average Confidence Drop (lower is better)} & 28.75\% & 37.08\% & 21.21\% & 15.43\%\\
\hline
\textbf{Prediction Change Percentage (lower is better)} & 35.33\% & 50.58\% & 27.05\% & 15.90\%\\
\hline
\end{tabular}}
\end{table*}

We retain the Ensemble-CAM regions by keeping their values intact. In the case of other CAMs (Grad-CAM, HiResCAM, and Grad-CAM++), we retain the top 10\% of regions by preserving their values and setting the rest to zero.
The goal is to retain regions of similar size for comparison. This step is designed to assess the impact of the most critical regions identified by the CAMs.  
Fig. \ref{fig:rtn} illustrates a comparison of the Ensemble-CAM explanation with that of other methods.

We compute the average confidence drop over the entire test dataset for all the CAM methods. The average confidence drop for Grad-CAM, HiResCAM, Grad-CAM++, and Ensemble-CAM is 28.75\%, 37.08\%, 21.21\%, and 15.43\%, respectively. This shows that the average confidence drop is the lowest for Ensemble-CAM.

The prediction change percentage for Grad-CAM, HiResCAM, Grad-CAM++, and Ensemble-CAM is 35.33\%, 50.58\%, 27.05\%, and 15.90\%, respectively. The prediction change percentage is the lowest for Ensemble-CAM, which validates the significance of Ensemble-CAM regions. The results are shown in Table \ref{tabel_rtn}.

\section{Discussion}\label{Discuss}
The Ensemble-CAM results demonstrate that the most relevant facial regions used by a deep learning-based face PAD model for making its decision can be localized more precisely and robustly by combining multiple interpretability techniques — Grad-CAM, HiResCAM, and Grad-CAM++. This ensemble approach integrates the strengths of each individual CAM variant to generate a more stable and informative heatmap, especially in the presence of presentation attacks. The explanations generated by Ensemble-CAM are compared to those of Grad-CAM, HiResCAM, and Grad-CAM++ using the retention method for evaluation. The evaluation results show that the Ensemble-CAM more specifically and accurately highlights the most significant regions (for the model's output) compared to Grad-CAM, HiResCAM, and Grad-CAM++. These regions (highlighted by Ensemble-CAM) reflect the discriminative features used by the model to distinguish between real and spoofed faces. 

While each constituent CAM method has its own merits — Grad-CAM being computationally efficient, Grad-CAM++ offering finer localization, and HiResCAM providing high-resolution focus — Ensemble-CAM combines them to improve robustness and reduce the sensitivity to noise or spurious activations. Although this ensemble method introduces moderate computational overhead due to the generation and fusion of multiple CAMs, its implementation remains manageable within the existing deep learning pipeline. The additional cost is justified by the enhanced consistency and precision of the visual explanations, particularly under challenging conditions such as fine-grained spoof patterns or ambiguous facial cues.

The Ensemble-CAM technique enhances the interpretability of the face PAD model by clearly revealing the regions responsible for detecting presentation attacks. This not only supports transparency in model decision-making but also fosters greater trust in the deployment of face PAD systems. Moreover, the insights derived from Ensemble-CAM visualizations can inform improvements in model training and spoof detection strategies, leading to better generalization and performance in real-world applications.

\section{Conclusion}\label{end}
This paper presents a novel visual explanation method, Ensemble-CAM, for deep learning-based face presentation attack detection. The Ensemble-CAM method integrates Grad-CAM, Grad-CAM++, and HiResCAM to enhance interpretability. The experimental results demonstrate that individual gradient-based CAM methods, such as Grad-CAM, HiResCAM, and Grad-CAM++, each have their own strengths and limitations in terms of localization precision, robustness, and sensitivity to gradients. Ensemble-CAM effectively addresses these issues by combining their complementary properties, resulting in more precise and reliable localization, especially under challenging or ambiguous attack scenarios. Ensemble-CAM offers more specific localization of the discriminative facial regions that contribute to the model's decision-making process.
This approach not only increases the transparency and trustworthiness of face PAD systems but also assists developers and researchers in diagnosing potential weaknesses or biases in the model.
Ultimately, Ensemble-CAM serves as a powerful interpretability tool for face PAD models, promoting informed deployment, enhancing end-user confidence, and potentially guiding improvements in model training and generalization.

One possible direction of future work is to apply the Ensemble-CAM method to other biometric modalities, such as fingerprint and iris. This method can be used to highlight the most significant features of spoofed fingerprint and iris images.


\bibliographystyle{plain}  
\bibliography{reference}   

\end{document}